\documentclass[acmtog,authorversion,screen]{acmart}

\usepackage{booktabs} 

\usepackage[ruled]{algorithm2e} 

\SetAlFnt{\small}
\SetAlCapFnt{\small}
\SetAlCapNameFnt{\small}
\SetAlCapHSkip{0pt}

\acmJournal{TOG}

\AtBeginDocument{%
  \providecommand\BibTeX{{%
    \normalfont B\kern-0.5em{\scshape i\kern-0.25em b}\kern-0.8em\TeX}}}

\usepackage{calc}
\usepackage{enumitem}
\usepackage{xcolor}
\usepackage{multirow}
\usepackage{makecell}

\definecolor{myOrange}{rgb}{1,0.5,0.13}
\definecolor{myRed}{rgb}{0.95,0.18,0.10}
\definecolor{myBlue}{rgb}{0.31,0.40,0.78}

\setlist{noitemsep, leftmargin=7mm, topsep=0pt}

\let\originalleft\left
\let\originalright\right
\renewcommand{\left}{\mathopen{}\mathclose\bgroup\originalleft}
\renewcommand{\right}{\aftergroup\egroup\originalright}

\DeclareMathOperator*{\argmin}{arg\,min}
\newcommand{\diff}{\mathop{}\!\mathrm{d}}
\let\originalpartial\partial
\renewcommand{\partial}{\mathop{}\!\mathrm{\originalpartial}}

\setcopyright{acmlicensed}
\copyrightyear{2024}
\acmYear{2024}
\acmDOI{XXXXXXX.XXXXXXX}

\acmConference[Conference acronym 'XX]{Make sure to enter the correct
  conference title from your rights confirmation emai}{June 03--05,
  2018}{Woodstock, NY}
\acmISBN{978-1-4503-XXXX-X/18/06}

\acmSubmissionID{1061}

\begin{document}

\title{Generic 3D Diffusion Adapter Using Controlled Multi-View Editing}

\author{Hansheng Chen}
\email{hanshengchen@stanford.edu}
\affiliation{%
  \institution{Stanford University}
  \city{Stanford}
  \state{California}
  \country{USA}
  \postcode{94305}
}

\author{Ruoxi Shi}
\affiliation{%
  \institution{UC San Diego}
  \state{California}
  \country{USA}
}

\author{Yulin Liu}
\affiliation{%
  \institution{UC San Diego}
  \state{California}
  \country{USA}
}

\author{Bokui Shen}
\affiliation{%
  \institution{Apparate Labs}
  \state{California}
  \country{USA}
}

\author{Jiayuan Gu}
\affiliation{%
  \institution{UC San Diego}
  \state{California}
  \country{USA}
}

\author{Gordon Wetzstein}
\affiliation{%
  \institution{Stanford University}
  \state{California}
  \country{USA}
}

\author{Hao Su}
\affiliation{%
  \institution{UC San Diego}
  \state{California}
  \country{USA}
}

\author{Leonidas Guibas}
\affiliation{%
  \institution{Stanford University}
  \state{California}
  \country{USA}
}


\renewcommand{\shortauthors}{Chen and Shi, et al.}

\begin{abstract}

Open-domain 3D object synthesis has been lagging behind image synthesis due to limited data and higher computational complexity. To bridge this gap, recent works have investigated multi-view diffusion but often fall short in either 3D consistency, visual quality, or efficiency. This paper proposes \emph{MVEdit}, which functions as a 3D counterpart of SDEdit, employing ancestral sampling to jointly denoise multi-view images and output high-quality textured meshes. Built on off-the-shelf 2D diffusion models, MVEdit achieves 3D consistency through a training-free \emph{3D Adapter}, which lifts the 2D views of the last timestep into a coherent 3D representation, then conditions the 2D views of the next timestep using rendered views, without uncompromising visual quality. With an inference time of only 2-5 minutes, this framework achieves better trade-off between quality and speed than score distillation. MVEdit is highly versatile and extendable, with a wide range of applications including text/image-to-3D generation, 3D-to-3D editing, and high-quality texture synthesis. In particular, evaluations demonstrate state-of-the-art performance in both image-to-3D and text-guided texture generation tasks. Additionally, we introduce a method for fine-tuning 2D latent diffusion models on small 3D datasets with limited resources, enabling fast low-resolution text-to-3D initialization.
\end{abstract}

\ccsdesc[500]{Computing methodologies~Computer graphics}
\ccsdesc[500]{Computing methodologies~Artificial intelligence}

\keywords{diffusion models, 3D generation and editing, texture synthesis, radiance fields, differentiable rendering}

\begin{teaserfigure}
  \vspace{-5.9ex}
  \begin{flushright}
  \large Demo \& code: \urlstyle{tt}\url{https://lakonik.github.io/mvedit}
  \end{flushright}
  \vspace{1.5ex}
  \begin{center}
    \includegraphics[width=0.88\linewidth]{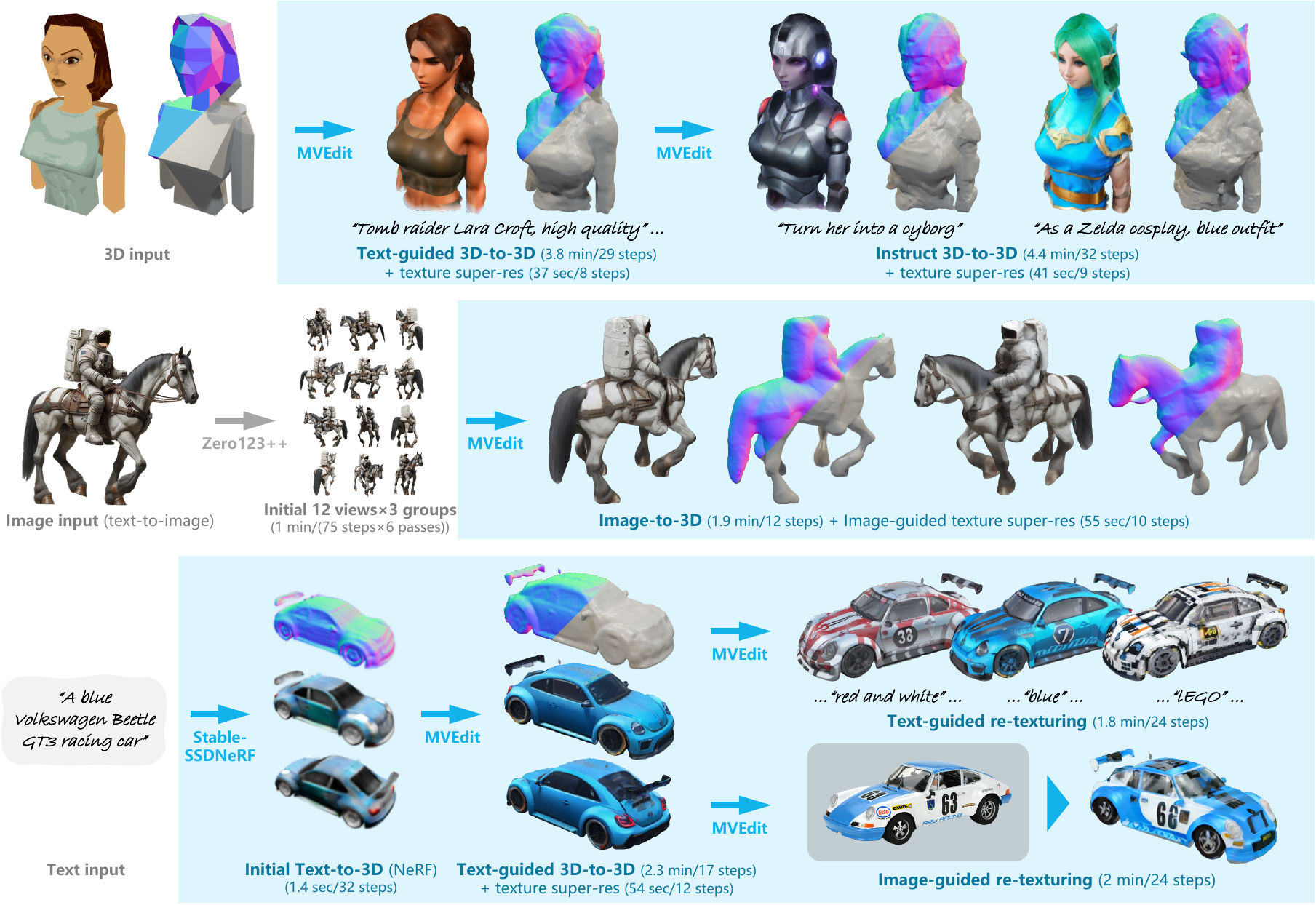}
    \caption{\textbf{Examples showcasing MVEdit's generality across various 3D tasks}, with associated inference times (on an RTX A6000) and the number of timesteps. For image-to-3D, note that the initial views by Zero123++ are not strictly 3D consistent (causing the failures in Fig.~\ref{fig:ablation_diff}), an issue remedied by MVEdit.}
    \label{fig:teaser}
  \end{center}
\end{teaserfigure}

\received{20 February 2007}
\received[revised]{12 March 2009}
\received[accepted]{5 June 2009}

\maketitle

\section{Introduction}

Data-driven 3D object synthesis in an open domain has gained wide research interest at the intersection of computer graphics and artificial intelligence.
Among the recent advances in generative modeling, diffusion models represent a significant leap in image generation and editing~\cite{controlnet,cfg,repaint,star_diffusion_models,ddpm,stablediffusion}. 
However, unlike 2D image models that benefit from massive datasets~\cite{laion5b} and a well-established grid representation, training a 3D-native diffusion model from scratch needs to grapple with the scarcity of large-scale datasets and the absence of a unified, neural-network-friendly representation, and has therefore been limited to closed domains or lower resolution~\cite{ssdnerf,rodin,lasdiffusion,functa,diffrf}. 

Multi-view diffusion has emerged as a promising approach to bridge the gap between 2D and 3D generation.
Yet, when adapting pretrained image diffusion models into multi-view generators, precise 3D consistency is not often guaranteed due to the absence of a 3D-aware model architecture. Score distillation sampling (SDS)~\cite{dreamfusion} further enforces 3D awareness by optimizing a neural radiance field (NeRF)~\cite{nerf} or mesh with multi-view diffusion priors, but they typically require hours-long optimization and often fall short in diversity and visual quality when compared to standard ancestral sampling (i.e., progressive denoising). 

To address these challenges, we present a generic solution for adapting pre-trained image diffusion models for 3D-aware diffusion under the ancestral sampling paradigm.
Inspired by ControlNet~\cite{controlnet}, we introduce the \emph{Controlled Multi-View Editing} (MVEdit) framework. Without fine-tuning, MVEdit simply extends the frozen base model by incorporating a novel training-free \emph{3D Adapter}. Inserted in between adjacent denoising steps, the 3D Adapter fuses multi-view 2D images into a coherent 3D representation, which in turn controls the subsequent 2D denoising steps without compromising image quality, thus enabling 3D-aware cross-view information exchange.

Analogous to the 2D SDEdit~\cite{sdedit}, MVEdit is a highly versatile 3D editor. Notably, when based on the popular Stable Diffusion image model~\cite{stablediffusion}, MVEdit can leverage a wealth of community modules to accomplish a diverse array of 3D synthesis tasks based on multi-modal inputs. 

Furthermore, MVEdit can utilize a real 3D-native generative model for geometry initialization. We therefore introduce StableSSDNeRF, a fast text-to-3D diffusion model fine-tuned from 2D Stable Diffusion, to complement MVEdit in high-quality domain-specific 3D generation.

To summarize, our main contributions are as follows:
\begin{itemize}
  \item We propose MVEdit, a generic framework for building 3D Adapters on top of image diffusion models, implementable on Stable Diffusion without the necessity for fine-tuning.
  \item Utilizing MVEdit, we develop a versatile 3D toolkit 
  and showcase its wide-ranging applicability in various 3D generation and editing tasks, as illustrated in Fig.~\ref{fig:teaser}.
  \item Additionally, we introduce StableSSDNeRF, a fast, easy-to-fine-tune text-to-3D diffusion model for initializing MVEdit.
\end{itemize}

\section{Related Work}

\subsection{3D-Native Diffusion Models}
\label{sec:related_3dnative}
We define 3D-native diffusion models as injecting noise directly into the 3D representations (or their latents) during the diffusion process. Early works~\cite{gaudi,functa} have explored training diffusion models on low-dimensional latent vectors of 3D representations, but are highly limited in model capacity. A more expressive approach is training diffusion models on triplane representations~\cite{eg3d}, which works reasonably well on closed-domain data~\cite{ssdnerf,triplanediff,3dgen,rodin}. Directly working on 3D grid representations is more challenging due to the cubic computation cost~\cite{diffrf}, so an improved multi-stage sparse volume diffusion model is proposed in \cite{lasdiffusion} and also adopted in \cite{one2345plus}. In general, 3D-native diffusion models face the challenge of limited data, and sometimes the extra cost of converting existing data to 3D representations (e.g., NeRF). These challenges are partially addressed by our proposed StableSSDNeRF (Section~\ref{sec:stablessdnerf}).

\subsection{Novel-/Multi-view Diffusion Models}
\label{sec:related_mvdiff}
Trained on multi-view images of 3D scenes, view diffusion models inject noise into the images (or their latents) and thus benefit from existing 2D diffusion research. \cite{3dim} have demonstrated the feasibility of training a conditioned novel view generative model using purely 2D architectures. Subsequent works~\cite{mvdream,zero123plus,zero123,wonder3d} achieve open-domain novel-/multi-view generation by fine-tuning the pre-trained 2D Stable Diffusion model~\cite{stablediffusion}. However, 3D consistency in these models is generally weak, as it is enforced only in a data-driven manner, lacking any inherent architectural bias.

To introduce 3D-awareness, \cite{renderdiffusion,dmv3d,forward} lift image features into 3D NeRF to render the denoised views. However, they are prone to blurriness due to the information loss during the 2D-3D-2D conversion. \cite{genvs,syncdreamer} propose 2D denoising networks conditioned on 3D projections, which generate crisp images but with slight 3D inconsistency. Inspired by the latter approach, MVEdit takes a significant step further by directly adopting pre-trained 2D diffusion models without fine-tuning, and enabling high-quality mesh output.

\subsection{Diffusion Models with 3D Optimization}
\label{sec:related_optim}
While the aforementioned approaches rely solely on feed-forward networks, optimization-based methods sometimes offer higher quality and greater flexibility, albeit at the cost of longer runtimes. \cite{dreamfusion} introduced the seminal Score Distillation Sampling (SDS), which optimizes a NeRF using a pretrained image diffusion model as a loss function. Some of its issues, such as limited resolution, the Janus problem, over-saturated colors, and mode-seeking behavior, have been addressed in subsequent works~\cite{dreamcraft3d,wang2023prolificdreamer,magic3d,fantasia3d,magic123}. Despite improvements, SDS and its variants remain time-consuming and often yield a degraded distribution compared to ancestral sampling. \cite{sparsefusion,instructnerf2nerf} alternate between ancestral sampling and optimization, which is also inefficient. A faster approach is seen in NerfDiff~\cite{nerfdiff}, which performs ancestral sampling only once and optimizes a NeRF within each timestep. However, if dealing with diverse open-domain objects, it would encounter the same blurriness issues due to NeRF disrupting the sampling process, a challenge to be addressed in this work.

\section{MVEdit: Controlled Multi-View Editing}

As discussed in Section~\ref{sec:related_mvdiff} and \ref{sec:related_optim}, although appending a 3D NeRF to the denoising network (Fig.~\ref{fig:arch_compare}~(b)) guarantees 3D consistency, it often leads to blurry results since NeRF typically averages the inconsistent multi-view inputs, resulting in inevitable loss. For latent diffusion models~\cite{stablediffusion}, the additional VAE decoding and encoding process can further exacerbate this issue. 

To address the 3D consistency challenge without interrupting the information flow from the input noisy view to the denoised view, we propose a new architecture containing a skip connection around the 3D model (Fig.~\ref{fig:arch_compare}~(c)) and its simplified version (Fig.~\ref{fig:arch_compare}~(d)). Based on the simplified architecture, we introduce the MVEdit framework shown in Fig.~\ref{fig:framework}, and provide a detailed elaboration below.

\subsection{Framework Overview}

\subsubsection{Preliminaries: SDEdit Using Single-Image Diffusion}
\begin{figure*}[t]
  \centering
  \includegraphics[width=\linewidth]{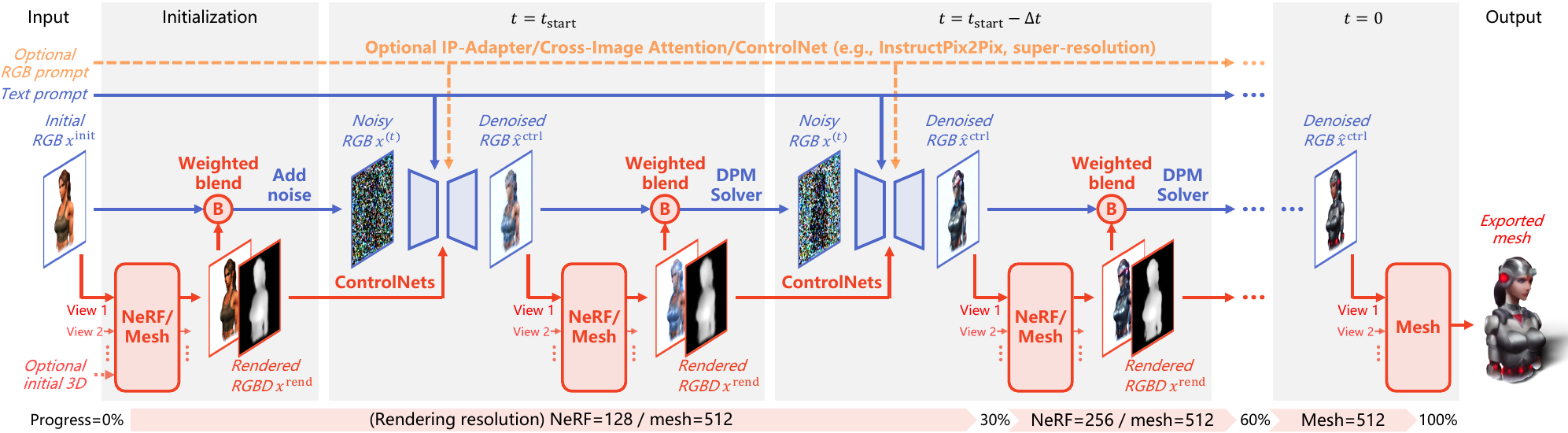}
  \caption{\textbf{The initialization and ancestral sampling process of MVEdit}. The original single-image SDEdit is shown in \textcolor{myBlue}{blue}, the additional 3D Adapter in \textcolor{myRed}{red}, and extra conditioning in \textcolor{myOrange}{orange}. For brevity, only the first view is depicted, and VAE encoding/decoding is omitted in cases involving latent diffusion.}
  \label{fig:framework}
\end{figure*}

Ignoring the \textcolor{myRed}{red} and \textcolor{myOrange}{orange} flow in Fig.~\ref{fig:framework}, the remaining \textcolor{myBlue}{blue} flow depicts the original SDEdit sampling process using the base text-to-image 2D diffusion model. For latent diffusion models, we omit the VAE encoding/decoding process for brevity. Given an initial RGB image $x^\text{init} \in \mathbb{R}^{C\times H\times W}$, SDEdit first perturbs the image with random noise $\epsilon \sim \mathcal{N}(0,I)$ following the Gaussian diffusion process:
\begin{equation}
    x^{(t)} = \alpha^{(t)} x^\text{init} + \sigma^{(t)} \epsilon,
\end{equation}
where $t \gets t_\text{start} \in [0, T]$ is a user-specified starting timestep, $\alpha^{(t)}, \sigma^{(t)}$ are scalars determined by the noise schedule, and $x^{(t)}$ denotes the noisy image. For the denoising step, the UNet $\hat{\epsilon}\left(x^{(t)},c,t\right)$ predicts the noise component $\hat{\epsilon}$ from the noisy image $x^{(t)}$, the condition $c$ (i.e., text prompt), and the timestep $t$. Afterwards, we can derive the denoised image $\hat{x}$ from the predicted noise $\hat{\epsilon}$:
\begin{equation}
    \hat{x} = \frac{\left(x^{(t)} - \sigma^{(t)} \hat{\epsilon}\left(x^{(t)},c,t\right)\right)}{\alpha^{(t)}}.
    \label{eq:x_pred}
\end{equation}
To move forward onto the next step, a generic diffusion ODE or SDE solver~\cite{song2021scorebased} can be applied to yield a less noisy image $x^{(t-\Delta t)}$ at a previous timestep $t-\Delta t$. In this paper, we adopt the DPMSolver++~\cite{dpmsolver}, and the solver step can be written as:
\begin{equation}
    x^{(t - \Delta t)} \gets \textit{DPMSolver}_z\left(\hat{x}, t, x^{(t)}\right),
    \label{eq:solver}
\end{equation}
where $z$ denotes the internal states of the solver. Recursive denoising can be executed by repeating Eq.~(\ref{eq:x_pred}) and Eq.~(\ref{eq:solver}) until reaching the denoised state $x^{(0)}$, thus completing the ancestral sampling process.

\begin{figure}[t]
  \centering
  \includegraphics[width=0.9\linewidth]{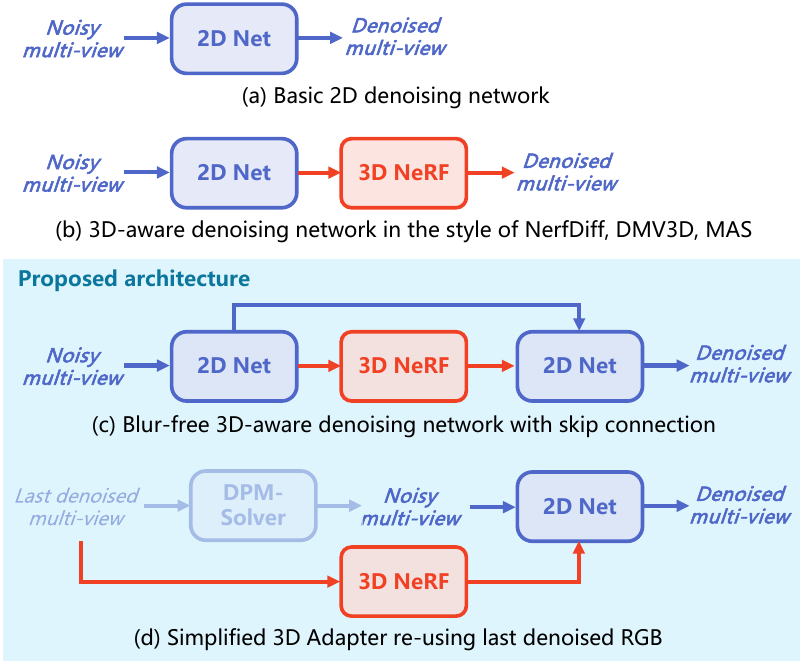}
  \caption{\textbf{Comparison among 3D-aware multi-view denoising architectures}. Adding skip connection around the 3D NeRF in (c) mitigates the potential blurriness issue in (b), but requires two 2D UNet passes within the same denoising timestep when extending the off-the-shelf 2D Stable Diffusion; our simplified architecture in (d) re-uses the denoised multi-view images from the last denoising timestep to reconstruct the 3D NeRF.}
  \label{fig:arch_compare}
\end{figure}

\begin{figure}[t]
  \centering
  \vspace{-0.5ex}
  \includegraphics[width=0.95\linewidth]{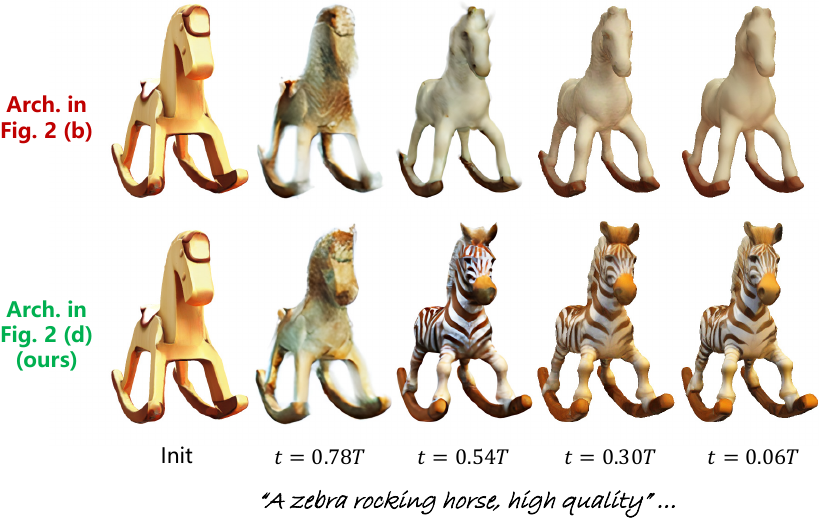}
  \vspace{-0.5ex}
  \caption{\textbf{Comparison between the two architectures}, based on the text-guided 3D-to-3D pipeline with $t^\text{start}=0.78T$. Rendered RGB images $x^\text{rend}_\text{RGB}$ across different timesteps are shown to visualize the sampling process.}
  \label{fig:ablation_ctrl}
\end{figure}

\subsubsection{MVEdit Using Multi-View Diffusion}
\label{sec:mvdiff}
In MVEdit, we adapt the single-image diffusion model into a 3D-consistent multi-view diffusion model via a novel 3D Adapter, depicted as the \textcolor{myRed}{red} flow in Fig.~\ref{fig:framework}. For each timestep, we first obtain the denoised images $\{\hat{x}_i\}$ of all the predefined views with known camera parameters $\{p_i\}$, where $i$ denotes the view index. Then, a 3D representation parameterized by $\phi$ can be reconstructed from these denoised views. In this paper, we employ optimization-based reconstruction approaches, using InstantNGP~\cite{ingp} for NeRF or DMTet~\cite{dmtet} for mesh. Thus, the 3D parameters $\hat{\phi}$ can be estimated by minimizing the rendering loss against the denoised images $\{\hat{x}_i\}$:
\begin{equation}
    \hat{\phi} = \argmin_{\phi}{\mathcal{L}_{\text{rend}}\left(\{\hat{x}_i, p_i\},\phi\right)}.
\end{equation}
Details on the loss and optimization will be described in Section~\ref{sec:nerf_mesh}. With the reconstructed 3D representation, a new set of images with RGBD channels $\{x_i^\text{rend}\}$ can be rendered from the views. These strictly 3D-consistent renderings are the results of multi-view aggregation, and tend to be blurry at early denoising steps. By feeding $x_i^\text{rend}$ to the ControlNets~\cite{controlnet} as a conditioning signal, a sharper image $\hat{x}_i^\text{ctrl}$ can be obtained via a second pass through the controlled UNet $\hat{\epsilon}^\text{ctrl}\left(x^{(t)},c_i,t,x^\text{rend}_i\right)$:
\begin{equation}
    \hat{x}^\text{ctrl}_i = \frac{\left(x^{(t)}_i - \sigma^{(t)} \hat{\epsilon}^\text{ctrl}\left(x^{(t)}_i,c_i,t,x^\text{rend}_i\right)\right)}{\alpha^{(t)}}.
    \label{eq:x_ctrl}
\end{equation}
Therefore, 3D-consistent sampling can be achieved by replacing $\hat{x}_i$ with $\hat{x}^\text{ctrl}_i$ in the solver step in Eq.~(\ref{eq:solver}). Eq.~(\ref{eq:x_ctrl}) effectively formulates the two-pass architecture shown in Fig.~\ref{fig:arch_compare}~(c), where the skip connection is essentially re-feeding the noisy multi-view into the second UNet. In practice, running two passes within a single denoising step appears redundant. Therefore, we use the rendered views from the last denoising step to condition the UNet of the next denoising step, which corresponds to the simplified architecture in Fig.~\ref{fig:arch_compare}~(d). 

Empirically, with Stable Diffusion~\cite{stablediffusion} as the base model, we find that off-the-shelf \emph{Tile} (conditioned on blurry RGB images) and \emph{Depth} (conditioned on depth maps) ControlNets can already handle RGB and depth conditioning for consistent multi-view generation, eliminating the necessity of training a custom ControlNet. However, recursive self-conditioning may amplify some unfavorable bias within Stable Diffusion, such as color drifting or over-sharpening/smoothing. Therefore, we adopt time-dependant dynamic ControlNet weights. Notably, we reduce the $Tile$ ControlNet weight when $t$ is large, otherwise the small denominator $\alpha^{(t)}$ in Eq.~(\ref{eq:x_ctrl}) at this time would significantly amplify any bias in the numerator. Reducing the ControlNet weight, however, leads to worse 3D consistency. To mitigate the consistency issue, we introduce an additional weighted blending operation for $t > 0.4 T$ only:
\begin{equation}
    \hat{x}^\text{blend}_i = w^{(t)} {x^\text{rend}_\text{RGB}}_i + (1 - w^{(t)}) \hat{x}^\text{ctrl}_i,
    \label{eq:blending}
\end{equation}
where ${x^\text{rend}_\text{RGB}}_i$ denotes the RGB channels of the rendered image, $\hat{x}^\text{ctrl}_i$ is the denoised image with reduced ControlNet weight, and $w^{(t)}$ is a time-dependant blending weight. The blended image $\hat{x}^\text{blend}_i$ is then treated as the denoised image to be fed into the DPMSolver.

\subsection{Robust NeRF/Mesh Optimization}
\label{sec:nerf_mesh}

The 3D Adapter faces the challenge of potentially inconsistent multi-view inputs, especially at the early denoising stage. Existing surface optimization approaches, such as NeuS~\cite{neus}, are not designed to address the inconsistency.
Therefore, we have developed various techniques for the robust optimization of InstantNGP NeRF~\cite{ingp} and DMTet mesh~\cite{dmtet}, using enhanced regularization and progressive resolution.

\subsubsection{Rendering}
For each NeRF optimization iteration, we randomly sample a 128$\times$128 image patch from all camera views. Unlike \cite{dreamfusion} that computes the normal from NeRF density gradients, we compute patch-wise normal maps from the rendered depth maps, which we find to be faster and more robust. 
For mesh rendering, we obtain the surface color by querying the same InstantNGP neural field used in NeRF.
For both NeRF and mesh, Lambertian shading is applied in the linear color space prior to tonemapping, with random point lights assigned to their respective views. 

\subsubsection{RGBA Losses} 
\label{sec:rgba_loss}
For both NeRF and mesh, we employ RGB and Alpha rendering losses to optimize the 3D parameters $\phi$ so that the rendered views $\{x^\text{rend}_i\}$ match the target denoised views $\{\hat{x}_i\}$. For RGB, we employ a combination of pixel-wise L\textsubscript{1} loss and patch-wise LPIPS loss~\cite{lpips}. 
For Alpha, we predict the target Alpha channel from $\{\hat{x}_i\}$ using an off-the-shelf background removal network~\cite{tracer} as in Magic123~\cite{magic123}. Additionally, we soften the predicted Alpha map using Gaussian blur to prevent NeRF from overfitting the initialization.

\subsubsection{Normal Losses}
\label{sec:normallosses}
To avoid bumpy surfaces, we apply an L\textsubscript{1.5} total variation (TV) regularization loss on the rendered normal maps:
\begin{equation}
    \mathcal{L}_\text{N} = \sum_{chw} \left\| w_{hw} \cdot \nabla_{hw} n^\text{rend}_{chw} \right\|^{1.5},
    \label{eq:normal_reg}
\end{equation}
where $n^\text{rend}_{chw} \in \mathbb{R}$ denotes the value of the $C\times H\times W$ normal map at index $(c, h, w)$, $\nabla_{hw} n^\text{rend}_{chw} \in \mathbb{R}^2$ is the gradient of the normal map w.r.t. $(h, w)$, and $w_{hw} \in [0, 1]$ is the value of a foreground mask with edge erosion. For image-to-3D, however, we can predict target normal maps from the initial RGB images $\{x^\text{init}_i\}$ using \cite{omnidata}, following \cite{dreamcraft3d}. In this case, we modify the regularization loss in Eq.~(\ref{eq:normal_reg}) into a normal regression loss:
\begin{equation}
    \mathcal{L}_\text{N} = \sum_{chw} \left\| w_{hw} \cdot \left(\nabla_{hw} n^\text{rend}_{chw} - \nabla_{hw} \hat{n}_{chw}\right)\right\|^{1.5},
\end{equation}
where $\hat{n}_{chw}$ denotes the value of the predicted normal map at index $(c, h, w)$. Additionally, we also employ a patch-wise LPIPS loss between the high-pass components of both the rendered and predicted normal maps, akin to the patch-wise RGB loss. 

\subsubsection{Ray Entropy Loss for NeRF}
To mitigate fuzzy NeRF geometry, we propose a novel ray entropy loss based on the probability of sample contribution. Unlike previous works~\cite{infonerf, latentnerf} that compute the entropy of opacity distribution or alpha map, we consider the ray density function:
\begin{equation}
    p(s) = T(s)\sigma(s),
\end{equation}
where $s$ denotes the distance, $\sigma(s)$ is the volumetric density and $T(s)=\exp{-\int_0^s \sigma(s) \diff{s}}$ is the ray transmittance. The integral of $p(s)$ equals the alpha value of the pixel, i.e., $a=\int_0^{+\inf}p(s)\diff{s}$, which is less than $1$. Therefore, the background probability is $1 - a$ and a corresponding correction term needs to be added when computing the continuous entropy of the ray as the loss function:
\begin{equation}
    \mathcal{L}_\text{ray} = \sum_r \int_{0}^{+\inf}{\negthickspace -p_r(s) \log{p_r(s)} \diff{s}} - 
    \underset{\raisebox{-4ex}{$\scriptstyle \text{background correction}$}}{\smash[b]{\underbrace{(1 - a_r) \log{\frac{1 - a_r}{d}}}_{}}},
\end{equation}
where $r$ is the ray index, and $d$ is a user-defined ``thickness'' of an imaginative background shell, which can be adjusted to balance foreground-to-background ratio.

\subsubsection{Mesh Smoothing Losses}
As per common practice, we employ the Laplacian smoothing loss~\cite{laplacian} and normal consistency loss to further regularize the mesh extracted from DMTet. 

\subsubsection{Implementation Details}

The weighted sum of the aforementioned loss functions is utilized to optimize the 3D representation. 
At each denoising step, we carry forward the 3D representation from the previous step and perform additional iterations of Adam~\cite{adam} optimization (96 for 3D or 48 for texture-only). During the ancestral sampling process, the rendering resolution progressively increases from 128 to 256, and finally to 512 when NeRF is converted into a mesh (for texture-only the resolution is consistently 512). When the rendering resolution is lower than the diffusion resolution 512, we employ RealESRGAN-small~\cite{realesrgan} for efficient super-resolution.

\section{MVEdit Applications and Pipelines}

\begin{figure}[t]
  \centering
  \includegraphics[width=0.9\linewidth]{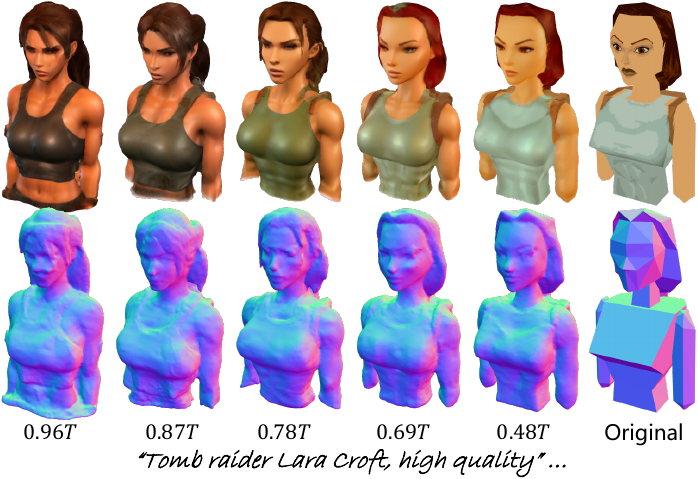}
  \caption{\textbf{Text-guided 3D-to-3D using the same seed but different $t^\text{start}$}. }
  \label{fig:t_sweep}
\end{figure}

In this section, we present details on various MVEdit pipelines. Their respective applications are showcased in Fig.~\ref{fig:teaser}, with details on inference times and the number of timesteps. Same as SDEdit, the initial timestep $t^\text{start}$ of these pipelines is adjustable, allowing control over the extent of editing, as shown in Fig.~\ref{fig:t_sweep}.

\subsection{3D Synthesis Pipelines}
3D synthesis pipelines, which fully utilize robust NeRF/mesh optimization techniques, begin with 32 views surrounding the object. These are progressively reduced to 9 views, helping to alleviate the computational cost of multi-view denoising at later stages. NeRF is always adopted as the initial 3D representation, with its density field converted into a DMTet mesh representation upon reaching 60\% completion. Various pipeline variants can then be constructed with unique input modalities and conditioning mechanisms.
\subsubsection{Text-Guided 3D-to-3D} Given an input 3D object, we randomly sample 32 surrounding cameras and render the initial multi-view images to initialize the NeRF. 
No additional modules are required, as Stable Diffusion is inherently conditioned on text prompts.
    
\subsubsection{Instruct 3D-to-3D} Inspired by Instruct-NeRF2NeRF~\cite{instructnerf2nerf}, we introduce the mesh-based Instruct 3D-to-3D pipeline. 
Extra image-conditioning is employed by feeding the initial multi-view images into an InstructPix2Pix ControlNet~\cite{ip2p,controlnet}.

\subsubsection{Image-to-3D} Using Zero123++~\cite{zero123plus} to generate initial multi-view images, MVEdit can lift these views into a high-quality mesh by resolving the initial 3D inconsistency. The original appearance can be preserved via image conditioning using IP-Adapter~\cite{ipadapter} and cross-image attention~\cite{zero123plus,crossimage}. Since Zero123++ can only generate a fixed set of 6 views, we augment the initialization by mirroring the input and repeating the generation process three times, yielding a total of 36 images. The pose of the input view can also be estimated using correspondences to the generated views, so that we have $36+1$ initial images in total. As the sampling process begins, this number is reduced to 32. 

\subsection{Re-Texturing Pipelines}
Given a frozen 3D mesh, MVEdit can generate high-quality textures from scratch (initialized with random Gaussian noise and $t_\text{start}=T$), or edit existing textures with a user-defined $t_\text{start}$. The number of views is scheduled to decrease from 32 to 7. This process is faster as it only requires optimizing the texture field. In this paper, we demonstrate basic re-texturing pipelines using text and image guidance (the latter using IP-Adapter and cross-image attention), while more pipelines can also be customized.

\subsection{Texture Super-Resolution Pipelines}
The texture super-resolution pipelines require only 6 views throughout the sampling process. We employ the \emph{Tile} ControlNet, originally trained for super-resolution, to condition the denoising UNet on the initial renderings. Consequently, the existing $Tile$ ControlNet in our 3D Adapter can be disabled to avoid redundancy. Additionally, image guidance can be implemented using cross-image attention, facilitating low-level detail transfer from a high-resolution guidance image. Adopting the SDE-DPMSolver++~\cite{dpmsolver}, these pipelines serve as a final boost to the 3D synthesis results.

\section{StableSSDNeRF: Fast Text-to-3D Initialization}
\label{sec:stablessdnerf}

Although text-to-3D generation is possible by chaining text-to-image and image-to-3D, we note that their ability in sculpting regular-shaped objects (e.g., cars) often lags behind 3D-native diffusion models trained specifically on category-level objects. 
However, as discussed in Section~\ref{sec:related_3dnative}, training 3D-native diffusion models often faces the challenges of limited data, making it difficult to complete creative tasks such as text-to-3D. To this end, we propose to fine-tune the text-to-image Stable Diffusion model into a text-to-triplane 3D diffusion model using the single-stage training paradigm in SSDNeRF~\cite{ssdnerf}, yielding the \emph{StableSSDNeRF}.

\begin{figure}[t]
  \centering
  \includegraphics[width=1.0\linewidth]{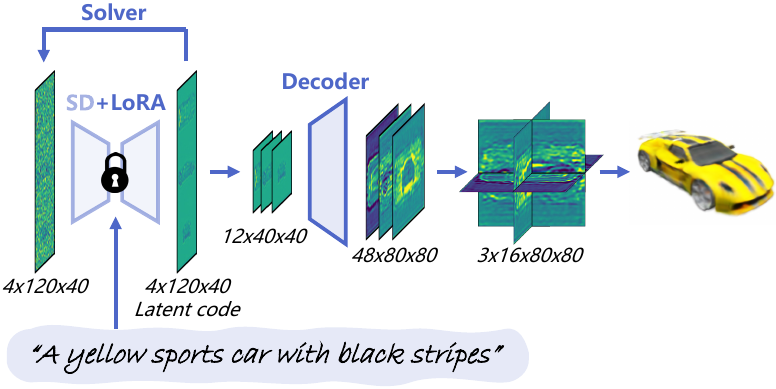}
  \caption{Architecture of StableSSDNeRF, consisting of a frozen Stable Diffusion UNet with LoRA fine-tuning, and a triplane latent decoder.}
  \label{fig:stablessdnerf}
\end{figure}

As shown in Fig.~\ref{fig:stablessdnerf}, StableSSDNeRF adopts a similar architecture to \cite{3dgen}, with a triplane latent diffusion model and a triplane latent decoder. However, instead of training a triplane VAE from scratch to obtain the triplane latents, we employ the off-the-shelf Stable Diffusion VAE encoder to obtain the image latents of orthographic views. These latents serve as the initial triplane latent for subsequent optimization, which aligns the triplane and image latent spaces initially, enabling the use of Stable Diffusion v2 as the backbone for triplane diffusion.

To fine-tune the model on 3D data, we adopt the LoRA approach \cite{lora} with a rank of 32 and freeze the base denoising UNet. Following the single-stage training of SSDNeRF, we jointly optimize the LoRA layers, the individual triplane latents, the triplane latent decoder (randomly initialized), and the triplane MLP layers. This optimization utilizes both the denoising mean-squared error (MSE) loss and the NeRF RGB rendering loss, the latter being a combination of pixel L\textsubscript{1} loss and patch LPIPS loss, as detailed in Section~\ref{sec:rgba_loss}. We fine-tune the model on the training split of ShapeNet-Cars~\cite{shapenet2015,srn} containing 2458 objects, with text prompts generated by BLIP~\cite{blip} and 128$\times$128 low resolution renderings. Using a batch size of 16 objects and 40k Adam iterations, training is completed in just 20 hours on two RTX3090 GPUs, making this approach particularly suitable for small-scale, domain-specific problems.

\section{Results and Evaluation}

\begin{table}[t]
  \caption{\textbf{Comparison on image-to-3D generation}. SyncDreamer and DreamCraft3D are not evaluated on the 248 objects due to slow inference.}
  \label{tab:img_to_3d}
  \scalebox{0.86}{%
  \setlength{\tabcolsep}{0.25em}
  \begin{tabular}{lccccccc}
    \toprule
    \multirow{2}[2]{*}{Method} & \multicolumn{3}{c}{248 GSO images} & \multicolumn{3}{c}{33 in-the-wild images} & \multirow{2}[2]{*}{\makecell{Infer. \\ time}} \\ 
    \cmidrule(lr){2-4}
    \cmidrule(lr){5-7}
    {} & LPIPS\textdownarrow & CLIP\textuparrow & FID\textdownarrow & {\footnotesize\makecell{Img-3D \\ Align.}}\textuparrow & {\footnotesize3D Plaus.}\textuparrow & {\footnotesize\makecell{Texture \\ Details}}\textuparrow \\
    \midrule
    SyncDreamer & - & - & - & 626 & 629 & 738 & > 20 min\\
    One-2-3-45 & 0.199 & 0.832 & 89.4 & 812 & 815 & 797 & 45 sec\\
    DreamGaussian & \underline{0.171} & 0.862 & 57.6 & 734 & 728 & 740 & 2 min\\
    Wonder3D & 0.240 & 0.871 & 55.7 & 848 & 903 & 829 & 3 min\\
    One-2-3-45++ & 0.219 & \underline{0.886} & \underline{42.1} & 1172 & 1177 & 1178 & 1 min\\
    DreamCraft3D & - & - & - & \underline{1189} & \underline{1202} & \underline{1210} & > 2 h\\
    Ours (MVEdit) & \textbf{0.139} & \textbf{0.914} & \textbf{29.3} & \textbf{1340} & \textbf{1339} & \textbf{1268} & 3.8 min \\
    \bottomrule
  \end{tabular}}
\end{table}

\begin{table}[t]
  \caption{\textbf{Comparison on text-guided texture generation}. *Our ablation study without skip connections resembles the method of TexFusion.}
  \label{tab:text2tex}
  \scalebox{0.9}{%
  \setlength{\tabcolsep}{0.3em}
  \begin{tabular}{lcccc}
    \toprule
    Methods & Aesthetic\textuparrow & CLIP\textuparrow & Infer. time & TV{\scriptsize$/10^7$}\\
    \midrule
    TEXTure & 4.66 & 25.39 & 2.0 min & 2.60 \\
    Text2Tex & \underline{4.72} & 24.44 & 11.2 min & 2.15 \\
    \midrule
    Ours (w/o skip, TexFusion)* & 4.68 & \textbf{26.34} & 1.5 min & 1.08 \\
    Ours (MVEdit) & \textbf{4.83} & \underline{26.12} & 1.6 min & 1.59 \\
    \bottomrule
  \end{tabular}}
\end{table}

\begin{figure*}[t]
  \centering
  \vspace{0.2ex}
  \includegraphics[width=\linewidth]{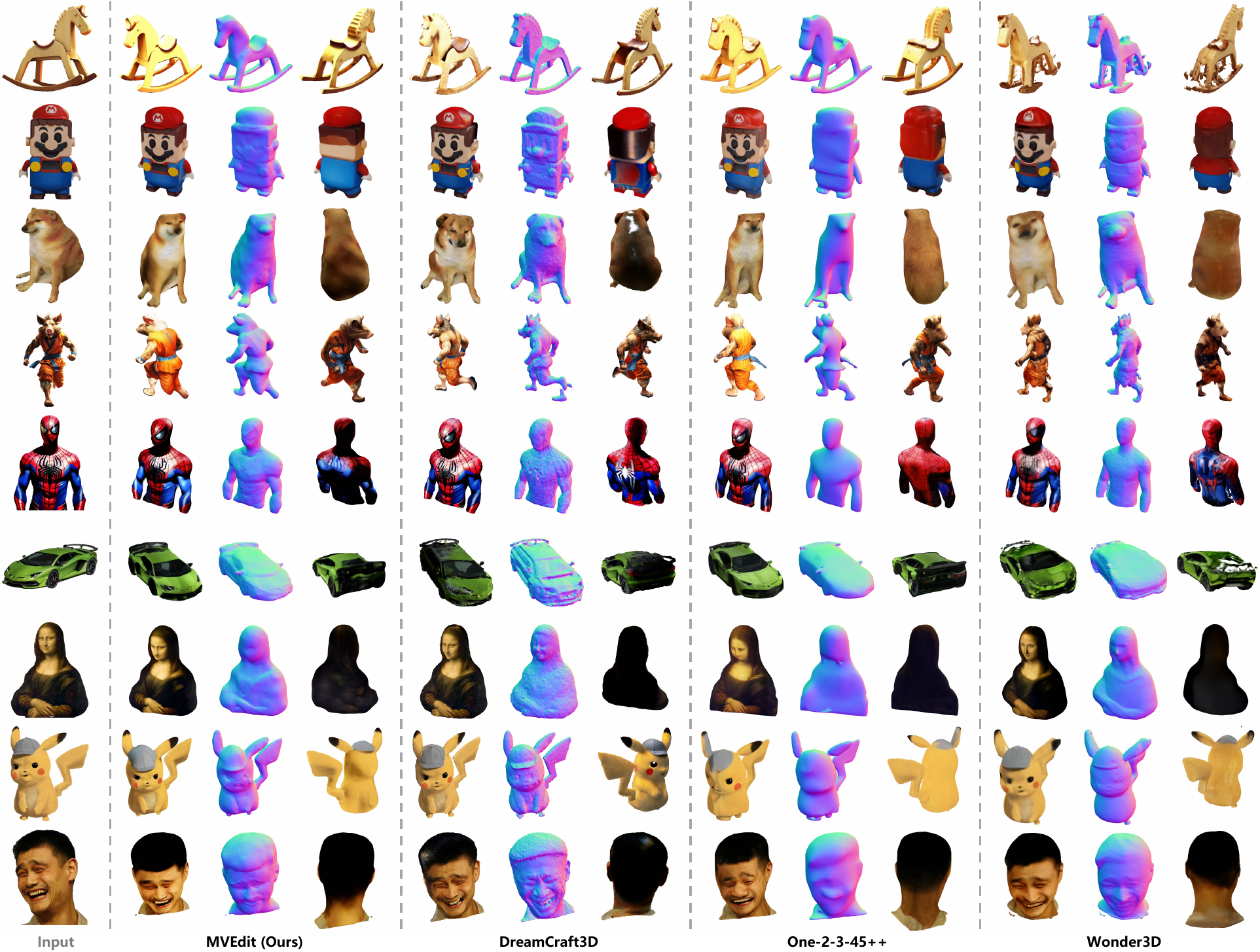}
  \caption{\textbf{Comparison of mesh-based image-to-3D methods on in-the-wild images}. Please zoom in for detailed viewing.}
  \label{fig:compare_imgto3d}
\end{figure*}

\subsection{Comparison on Image-to-3D Generation}
We compare the image-to-3D results of our MVEdit against those from previous state-of-the-art image-to-3D mesh generators, utilizing two test sets: 248 rendered images of objects sampled from the GSO dataset~\cite{gso}, and 33 in-the-wild images, which include demo images from prior studies, AI-generated images, and images sourced from the Internet. To evaluate the quality of the generated textured meshes, we render them from novel views and calculate quality metrics for these renderings. For the GSO test set, we calculate the LPIPS scores~\cite{lpips}, CLIP similarities~\cite{clip}, and FID scores~\cite{FID}, comparing the renderings of the generated meshes against the ground truth meshes. For the in-the-wild images without ground truths, we follow \cite{gpteval3d} and ask GPT-4V~\cite{openai2023gpt4} to compare the multi-view renderings from difference methods based on Image-3D Alignment, 3D Plausibility, and Texture Details. These comparisons allow us to compute the Elo scores~\cite{elo1967} of the evaluated methods, providing an automated alternative to costly user studies.

In Table~\ref{tab:img_to_3d}, we present the results for One-2-3-45~\cite{one2345}, DreamGaussian~\cite{dreamgaussian}, Wonder3D~\cite{wonder3d}, One-2-3-45++\cite{one2345plus}, and our own MVEdit (incorporating both image-to-3D and texture super-resolution) on the GSO test set. This comparison shows that MVEdit significantly outperforms the other methods on all metrics, while still offering a reasonable runtime. For the in-the-wild images, we extend our comparison to include SyncDreamer\cite{syncdreamer} and DreamCraft3D~\cite{dreamcraft3d}. Here, GPT-4V shows a distinct preference for our method, with MVEdit achieving Elo scores that exceed those of the SDS method DreamCraft3D, despite the latter's extensive object generation time of over two hours.

Fig.~\ref{fig:compare_imgto3d} further presents qualitative comparison among the top competitors. Wonder3D~\cite{wonder3d} generates multi-view images and normal maps for InstantNGP-based surface optimization, which can lead to broken structures due to multi-view inconsistency. One-2-3-45++\cite{one2345plus} utilizes the same multi-view generator as ours (i.e., Zero123++) but employs a multi-view-conditioned 3D-native diffusion model to generate signed distance functions (SDF) for surface extraction, yet this results in overly smooth surfaces with occasional missing parts. DreamCraft3D\cite{dreamcraft3d}, while capable of producing impressive geometric details through its hours-long distillation, generally yields noisy geometry and textures, sometimes even strong artifacts and the Janus problem. In contrast, our approach, while less detailed in geometry compared to SDS, is generally more robust and exhibits fewer artifacts or failures. This results in renderings that are visually more pleasing.

\subsection{Comparison on Text-Guided Texture Generation}

\begin{figure*}[t]
  \centering
  \includegraphics[width=1.0\linewidth]{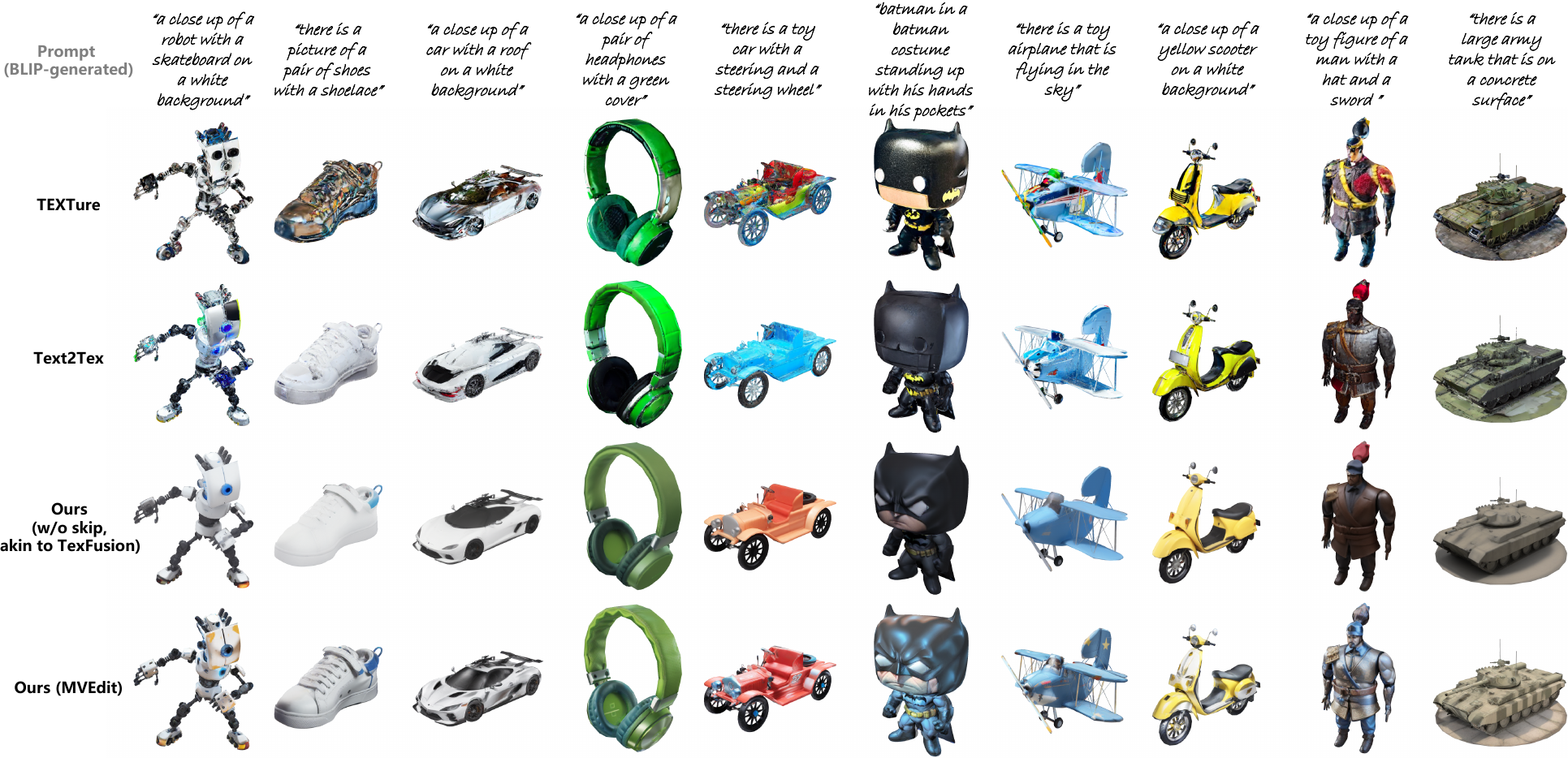}
  \caption{\textbf{Comparison on text-guided texture generation}. Please zoom in for detailed viewing. Note that the BLIP-generated text prompts may not accurately reflect the actual geometry, so it is impossible to generate texture maps that align perfectly with the prompts.}
  \label{fig:compare_text2tex}
\end{figure*}

We randomly select 92 objects from a high-quality subset of Objaverse~\cite{objaverse} and employed BLIP~\cite{blip} to generate text prompts from their rendered images. Using these textureless meshes and the generated prompts of these objects, we evaluate our MVEdit re-texturing pipeline against TEXTure~\cite{texture} and Text2Tex~\cite{text2tex}. TexFusion~\cite{texfusion} is not directly compared due to the unavailability of official code, but it closely resembles a scenario in our ablation studies, which will be discussed in Section~\ref{sec:arch}. We assess the quality of the generated textured meshes through rendered images, calculating Aesthetic~\cite{laion5b} and CLIP~\cite{clip,dreamfields} scores as the metrics. It is important to note, as shown in a user study by \cite{gpteval3d}, that Aesthetic scores more closely align with human preferences for texture details, whereas CLIP scores are less sensitive. Table~\ref{tab:text2tex} shows that MVEdit outperforms TEXTure and Text2Tex in both metrics by a clear margin and does so with greater speed. 

Fig.~\ref{fig:compare_text2tex} presents a quantitative comparison among the tested methods. Both TEXTure and Text2Tex generate slightly over-saturated colors and produce noisy artifacts. In contrast, MVEdit produces clean, detailed textures with a photorealistic appearance and strong text-image alignment.

\subsection{Ablation Studies}

\begin{figure}[t]
  \centering
  \includegraphics[width=0.89\linewidth]{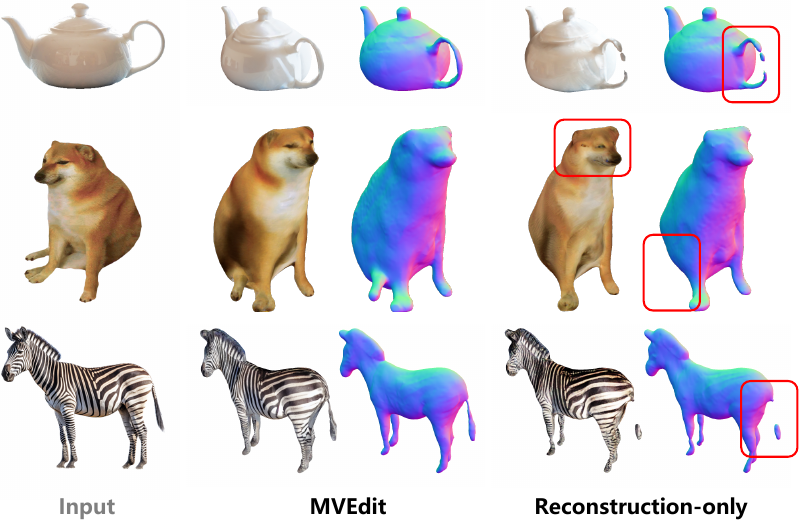}
  \caption{\textbf{Ablation study on the effectiveness of MVEdit in resolving multi-view inconsistency}. Without MVEdit diffusion, the reconstruction-only approach leads to broken thin structures and ambiguous textures.}
  \label{fig:ablation_diff}
\end{figure}

\begin{table}[t]
  \caption{\textbf{Quantitative ablation study on the effectiveness of MVEdit in resolving multi-view inconsistency}.}
  \label{tab:ablation_diff}
  \scalebox{0.9}{%
  \setlength{\tabcolsep}{0.5em}
  \begin{tabular}{lccccc}
    \toprule
    Methods & \makecell{Img-3D \\ Align.}\textuparrow & 3D Plaus.\textuparrow & \makecell{Texture \\ Details}\textuparrow \\
    \midrule
    Ours (MVEdit) & 1340 & 1339 & 1268 \\
    Ours (Reconstruction-only) & 1275 & 1252 & 1241 \\
    \bottomrule
  \end{tabular}}
\end{table}

\begin{figure*}[t]
  \centering
  \includegraphics[width=1.0\linewidth]{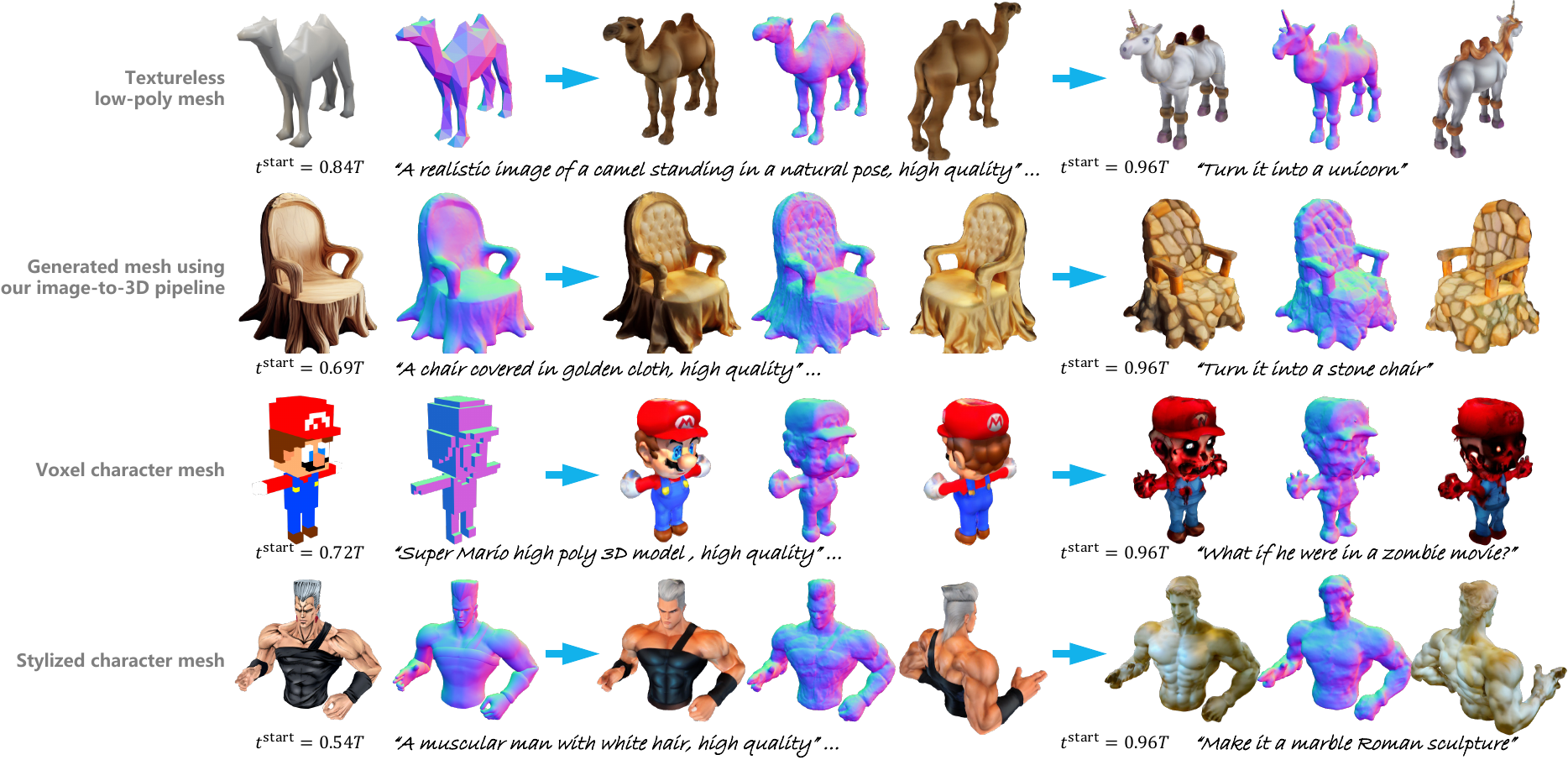}
  \caption{\textbf{Results of our text-guided 3D-to-3D and instruct 3D-to-3D pipelines}.
  }
  \label{fig:3dto3d}
\end{figure*}

\subsubsection{Effectiveness of the 3D Adapter with a Skip Connection}
\label{sec:arch}
To validate the effectiveness of our ControlNet-based 3D Adapter, we conduct an ablation study by removing the ControlNet, and set the blending weight $w^{(t)}$ in Eq.~\ref{eq:blending} to 1 for all timesteps, effectively constructing an architecture without a skip connection, as shown in Fig.~\ref{fig:arch_compare}~(b). For text-guided texture generation, sampling without skip connections is fundamentally akin to TexFusion~\cite{texfusion}, which is known to yield textures with fewer details due to the information loss. This is confirmed by our quantitative results presented in Table~\ref{tab:text2tex}, which show a notable decrease in the Aesthetic score and Total Variation. Qualitative comparisons in Fig.~\ref{fig:compare_text2tex} further illustrate the visual gap between the two architectures. For 3D-to-3D editing, Fig.~\ref{fig:ablation_ctrl} shows that the skip connection plays a crucial role not only in producing crisp textures but also in enhancing geometric details (e.g., the ears and knees of the zebra).

\subsubsection{Image-to-3D: MVEdit v.s. Reconstruction-Only}
To validate that our image-to-3D pipeline effectively resolves the 3D inconsistency in the initial views generated by Zero123++, we conduct an ablation study by using only the initial views for robust NeRF/mesh optimization, thus bypassing the denoising UNet/DPMSolver and leaving only the reconstruction side. Quantitatively, the GPT-4V evaluation results in Table~\ref{tab:ablation_diff} reveal a clear gap between MVEdit and the reconstruction-only method, underscoring MVEdit's effectiveness. 
Qualitatively, as observed in Fig.\ref{fig:ablation_diff}, the reconstruction-only method tends to result in broken thin structures and less defined textures, a common consequence of multi-view misalignment.

\begin{figure}[t]
  \centering
  \includegraphics[width=1.0\linewidth]{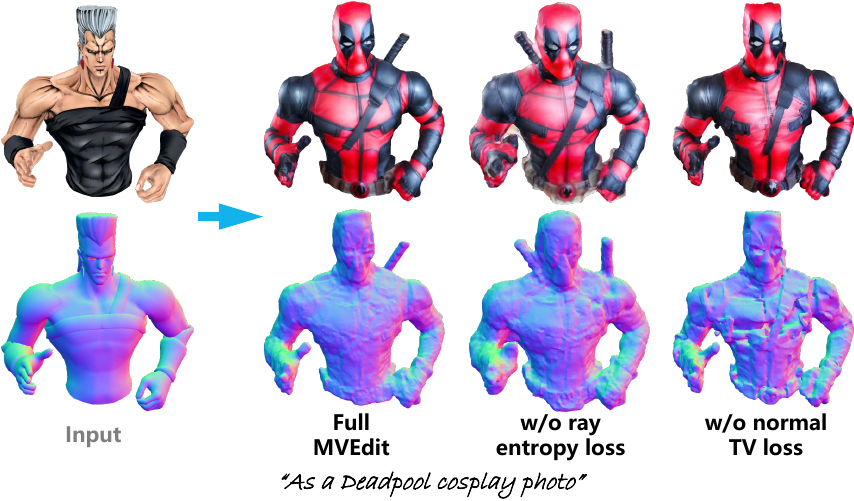}
  \caption{\textbf{Ablation study on the regularization loss functions}, based on the instruct 3D-to-3D pipeline with $t^\text{start} = 1.0T$, using the same seed.}
  \label{fig:ablation_losses}
\end{figure}

\begin{figure*}[t]
  \centering
  \includegraphics[width=1.0\linewidth]{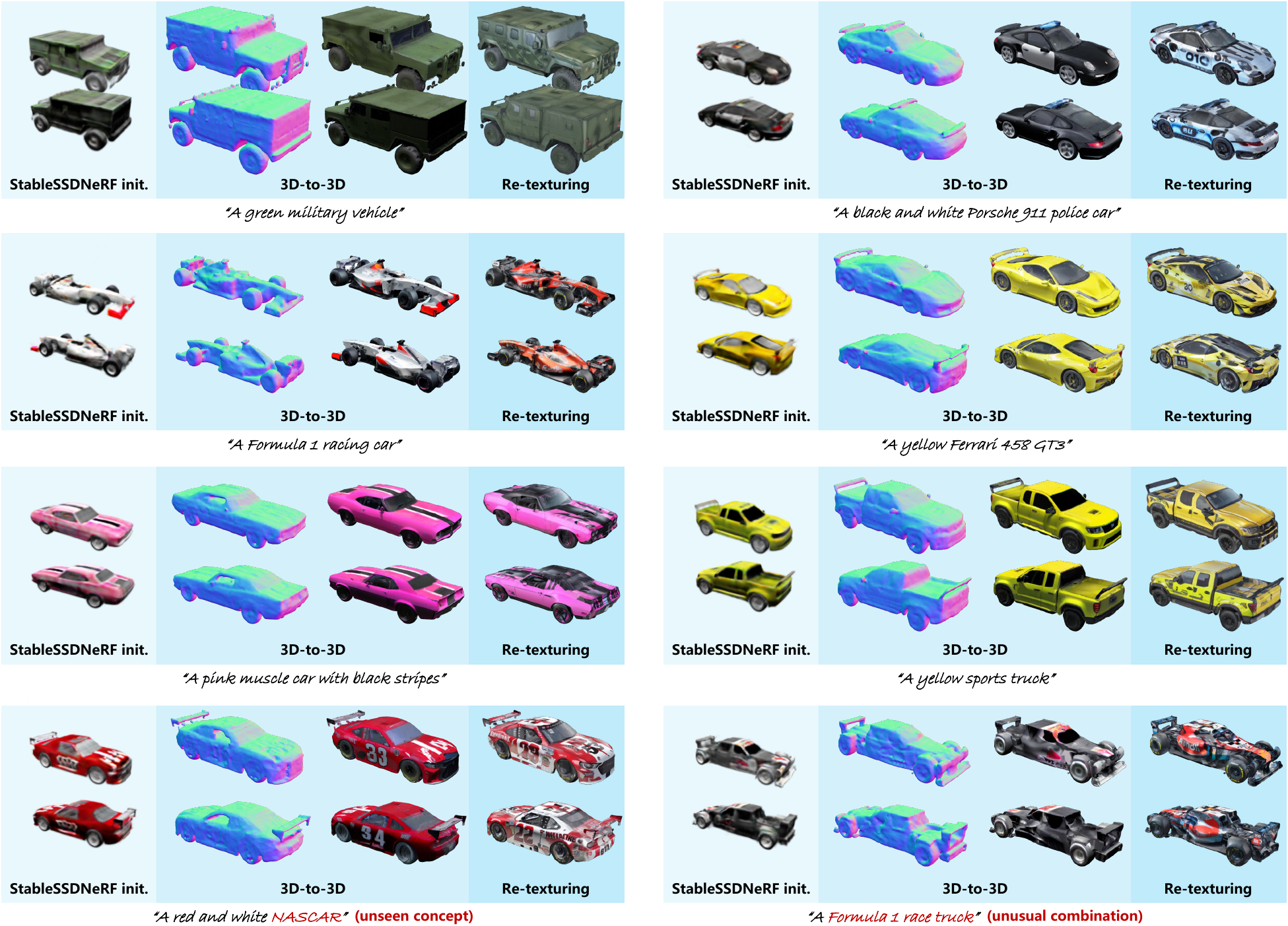}
  \caption{\textbf{Results of text-to-3D generation using StableSSDNeRF and MVEdit pipelines}.}
  \label{fig:text-to-3D}
\end{figure*}

\subsubsection{Effectiveness of the Regularization Loss Functions}
In Fig.~\ref{fig:ablation_losses}, we showcase the results of instruct 3D-to-3D editing under three settings: the full MVEdit, the one without ray entropy loss, and the one without normal TV loss. It can be seen that: removing the ray entropy loss results in inflated geometry and less defined textures, a consequence of initializing DMTet with a fuzzy density field; removing the normal TV loss appears to have little impact on texture quality but leads to numerous holes in the geometry. Although the degradation in quality from these ablations is apparent to humans, especially when viewed interactively in 3D, we note that existing metrics, including Aesthetic score, CLIP score, and even the GPT-4V metrics, struggle to capture these differences. Therefore, we do not include quantitative evaluations for these ablation studies.

\subsection{3D-to-3D Editing Results and Discussions}
In Fig.~\ref{fig:3dto3d}, we showcase results from both the text-guided 3D-to-3D pipeline and the instruct 3D-to-3D pipeline (with texture super-resolution), edited from four types of inputs: a textureless low-poly mesh, a mesh generated by our image-to-3D pipeline, a voxel character mesh, and a stylized character mesh. As demonstrated in the figure, all inputs are adeptly handled, resulting in prompt-accurate appearances, intricate textures, and detailed geometry, thereby highlighting the versatility of our 3D-to-3D pipelines.

\subsection{Text-to-3D Generation Results and Discussions}
In Fig.~\ref{fig:text-to-3D}, we showcase results of text-to-3D generation using a combination of StableSSDNeRF and MVEdit pipelines. Thanks to the knowledge transfer from a large image diffusion model, StableSSDNeRF is able to follow never-seen prompts despite being fine-tuned only on low-resolution renderings of 2458 ShapeNet 3D Cars, generating the correct combination of colors and style. Notably, it can even generalize to completely unseen concept (\emph{NASCAR}), or to unusual combinations (\emph{Formula 1} and \emph{truck}). When further processed using the text-guided 3D-to-3D and re-texturing pipelines, conditioned on the same input prompts, our method successfully produces diverse, high-quality, photorealistic cars within just 4 minutes.

\subsection{Sample Diversity}
\begin{figure}[t]
  \centering
  \includegraphics[width=1.0\linewidth]{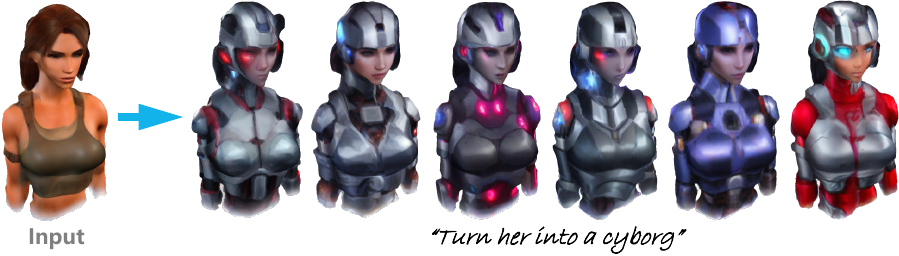}
  \caption{\textbf{An example showcasing the diversity of the generated the samples}, based on the instruct 3D-to-3D pipeline with $t^\text{start} = 1.0T$.}
  \label{fig:sample_diversity}
\end{figure}

Unlike SDS approaches that exhibit a mode-seeking behavior, MVEdit can generate variations from the exact same input using different random seeds. An example is shown in Fig.~\ref{fig:sample_diversity}.

\section{Conclusion and Limitations}

In this work, we have bridged the gap between 2D and 3D content creation with the introduction of MVEdit, a generic approach for adapting 2D diffusion models into 3D diffusion pipelines. Our novel training-free 3D Adapter, leveraging off-the-shelf ControlNets and a robust NeRF/mesh optimization scheme, effectively addresses the challenge of achieving 3D-consistent multi-view ancestral sampling while generating sharp details. Additionally, we have developed StableSSDNeRF for domain-specific 3D initialization. Extensive quantitative and qualitative evaluations across a range of tasks have validated the effectiveness of the 3D Adapter design and the versatility of the associated pipelines, showcasing state-of-the-art performance in both image-to-3D and texture generation tasks.

Despite the achievements, the MVEdit 3D-to-3D pipelines still face the Janus problem when $t^\text{start}$ is close to $T$, unless controlled explicitly by directional text/image prompts. Furthermore, the off-the-shelf ControlNets, not being originally trained for our task, can introduce minor inconsistencies and sometimes impose their own biases. Future work could train improved 3D Adapters for strictly consistent and Janus-free multi-view ancestral sampling.

\section{Acknowledgements}
This project was in part supported by Vannevar Bush Faculty
Fellowship, ARL grant W911NF-21-2-0104, Google, and Samsung.
We thank the members of Geometric Computation Group, Stanford Computational Imaging Lab, and SU Lab for useful feedback and discussions. Special thanks to Yinghao Xu for sharing the data, code, and results for image-to-3D evaluation.

\bibliographystyle{ACM-Reference-Format}
\bibliography{main}


\end{document}